\documentclass[a4paper,12pt]{article}

\usepackage[utf8]{inputenc}
\usepackage[OT1]{fontenc}

\usepackage[pdftex]{graphicx}
\usepackage[english]{babel}

\usepackage{amsmath}
\usepackage{amsfonts}
\usepackage{amssymb}
\usepackage{amsthm}

\usepackage{multicol}
\usepackage{xcolor}
\usepackage[pdftex]{hyperref}

\newcommand{\beq}{\begin{equation}}
\newcommand{\eeq}{\end{equation}}
\newcommand{\beqna}{\begin{eqnarray}}
\newcommand{\eeqna}{\end{eqnarray}}
\newcommand{\bit}{\begin{itemize}}
\newcommand{\eit}{\end{itemize}}

\begin{document}
 \title{Robust Computational Extraction of Non-Enhancing Hypercellular Tumor Regions from Clinical Imaging Data}
 \author{$^1$A. Brawanski, $^2$Th. Schaffer, $^2$F. Raab, $^1$K.-M. Schebesch, \\
 $^1$M. Schrey, $^1$Chr. Doenitz, $^3$A. M. Tom\'e and $^2$E. W. Lang, \\
 $^1$Neurosurgery, University Hospital, 93053 Regensburg, Germany \\
 $^2$CIML, Biophysics, University of Regensburg, 93053 Regensburg, Germany \\
 $^3$IEETA, DETI, Universidade de Aveiro, 3810-187 Aveiro, Portugal}

 \maketitle

 \begin{abstract}
Accurate identification of non-enhancing hypercellular (NEH) tumor regions is an unmet need in neuro-oncological imaging, with significant implications for patient management and treatment planning. We present a robust computational framework that generates probability maps of NEH regions from routine MRI data, leveraging multiple network architectures to
address the inherent variability and lack of clear imaging boundaries. Our approach was validated against independent clinical markers - relative cerebral blood volume (rCBV) and enhancing tumor recurrence location (ETRL) - demonstrating both methodological robustness and biological relevance. This framework enables reliable, non-invasive mapping of NEH tumor compartments, supporting their integration as imaging biomarkers in clinical workflows and advancing precision oncology for brain tumor patients.
 \end{abstract}

\section{Introduction}

Non-enhancing hypercellular (NEH) tumor regions are increasingly recognized as clinically significant in neuro-oncology, particularly for their role in disease progression and treatment resistance \cite{Cluceru2020},\cite{Brenner2022}, \cite{Bobholz2024}. However, NEH regions are difficult to delineate using standard imaging techniques due to their poorly defined boundaries and overlap with surrounding edema \cite{Bobholz2024a}, \cite{Familiar2024}. This lack of standardized, reproducible methods for NEH identification limits their integration as imaging biomarkers in clinical practice.

To establish the clinical and biological relevance of the extracted NEH regions, we validated our results against independent markers such as relative cerebral blood volume (rCBV) and the spatial pattern of tumor recurrence (ETRL), both of which are known to correlate with hypercellular tumor components \cite{Provenzale2006}, \cite{Price2011}, \cite{Hu2009}.
To assess whether the observed spatial metrics were significantly greater than expected by chance, a non-parametric permutation test was performed for each metric. This involved randomly flipping the sign of deviations from a null baseline (e.g., 0 for distances, 1 for ratios) 10,000 times and recalculating the mean value for each permutation. This generated a
null distribution for each metric under the assumption of no systematic spatial relationship. The observed mean was then compared to this distribution to compute both one-sided and two-sided p-values. Our goal is to provide a robust, non-invasive tool for NEH identification that can be integrated into clinical workflows like planning tumor resection or radiation therapy, supporting precision imaging and improved patient management in neuro-oncology.

Recognizing the challenges posed by the absence of a definitive ground truth and the inherent variability in NEH definition, we now present an extended approach that generates probability maps based on multiple network architectures. The current study aims to capture the most likely distribution of NEH regions, rather than enforcing rigid boundaries.
\subsection{Recent related own work}

In a recent study we developed an approach to extract information about non-enhancing hypercellular (NEH) tumor regions from curated MRI data \cite{Schaffer2025}. Utilizing datasets from the BraTS 2018 \cite{Bakas2019} and BraTS 2021 \cite{Baid2021} challenges, we leveraged differences in the segmentation of the necrosis label to delineate the NEH compartment. Specifically, NEH regions were reconstrcted by applying a BraTS 2021-trained model to BraTS 2018 data and identifying necrosis regions located outside the enhancing tumor areas, based on prediction differences and subsequent morphological filtering of the segmentation masks.

Based on these results, we created a new dataset derived from the BraTS 2021 pseudo-ground truth, now including an explicit NEH segment. This dataset comprises $1,536$ patients and provides four segmentation classes: enhancing tumor (ET), necrosis (NC), edema (ED), and non-enhancing hypercellular tumor (NEH).

To enable generalization to unseen cases, we trained a modified 3D U-Net architecture, called PAUNet, on this new dataset. PAUNet features a high-resolution decoding branch and filtered skip connections, specifically optimized for accurate segmentation of small, non-enhancing tumor compartments in brain MRI \cite{Schaffer2025a}. The resulting model yielded four distinct segments deduced from the standard four MRI sequences. It outperformed the BraTS challenge winner models from 2018 to 2020 and achieved Dice - S\o{}rensen scores close to the best models from the BraTS 2021 competition.
\section{Methods}

To generate probability maps and further improve segmentation robustness, in this study we additionally trained a network with a topology different from PAUNet. We selected UNETR++ \cite{Shaker2022}, an advanced transformer-based model that builds upon UNETR (“U-Net with Transformer Encoder”) \cite{Hatamizadeh2021}. Unlike traditional U-Net architectures that use convolutional encoders, UNETR employs a pure transformer encoder to globally model volumetric relationships, which can be advantageous for segmenting irregular and small regions such as NEH. Also, UNETR++ further enhances this architecture with improved multi-scale skip connections and more  training, enabling better performance on fine structures and boundaries.

This dual-network approach, combining PAUNet \cite{Schaffer2025} and UNETR++, allowed us to generate robust probability maps for NEH segmentation by leveraging the complementary strengths of convolutional and transformer-based architectures. This is particularly important given the known challenges of distinguishing NEH from surrounding edema and the inherent variability in segmenting poorly defined tumor compartments \cite{Xie2005}.
\subsection{Extraction and estimation of NEH regions}

We used the publicly available UNETR++ implementation from Shaker’s GitHub repository $(https://github.com/Amshaker/unetr_plus_plus)$, which is embedded within the Heidelberg nnU-Net framework based on nnU-Net version 1 \cite{Isensee2021}. No changes were made to the network topology but we adapted the configuration to output four segments (adding NEH to the original three) and cropped the data to match the dimensions of the BraTS 2018 dataset, for which the model was originally tailored.

For training, we used $1,212$ cases from the new dataset with four segments, splitting them into $969$ training and $243$ validation cases. The network was trained for $595$ epochs, with no further improvement in validation loss observed for over two hours, at which point training was stopped (no additional “best model” was saved). All training was conducted using PyTorch on an NVIDIA RTX 4090 GPU, with a mean calculation time of approximately $80$ seconds per epoch.

Final NEH probability maps were generated by combining outputs from the PAUNet and UNETR++ models. First, voxel-wise probabilities from both networks were averaged to produce a consensus likelihood map, leveraging their complementary strengths in capturing local (PAUNet) and global (UNETR++) spatial features. To reduce noise and ensure spatial coherence, a 3D Gaussian filter with a standard deviation of $\sigma = 1.5\ mm$ was applied to the mean probability map. This step smoothed isolated high-probability voxels while preserving contiguous regions likely to represent true NEH components. Finally, the filtered probability map was thresholded at $p = 0.5$ to generate binary masks suitable for clinical interpretation, balancing sensitivity and specificity in delineating NEH boundaries. This yielded two segments in the final map, one label with very high probability and one with low probability.

This new dataset was used to train an additional UNETR-staple++ \cite{Krishnan2023} model, configured to segment only two labels. No changes were made to the original architecture except adjusting the number of input and output channels $(n=2)$. The network was trained for the full $1,000$ epochs as recommended in the default configuration by Isensee et al. \cite{Isensee2021}(13).

With this setup, we generated three distinct models:

\bit
\item
PAUNet - producing four segments,
\item
UNETR++ - producing four segments,
\item
UNETR-staple++ - producing two segments.
\eit

Segmentation performance was quantified using the Dice-S\o{}rensen similarity coefficient (overlap), Hausdorff95 distance (boundary agreement), Jacquard index (intersection-over-union), and Surface Dice at $2\ mm$ tolerance (clinical contouring relevance).
\subsection{Clinical validation of segmentation models}
For clinical validation of these three models we used two datasets, namely the University of Pennsylvania Glioblastoma (UPenn-GBM) dataset (https://www.cancerimagingarchive.net/collection/upenn-gbm/) \cite{Bakas2022} and the Rio Hortega University Hospital (RHUH) glioblastoma dataset (https://www.cancerimagingarchive.net/collection/rhuh-gbm/) \cite{Cepeda2023}. Each of them provides additional information that was not utilized in training the models.

Clinical validation is based on the following hypotheses:
\bit
\item
H1: The segmented NEH regions are expected to demonstrate higher absolute pseudo-relativ cerebral blood volume (ap-rCBV) values than the segmented edema regions, but lower values than the contrast-enhancing tumor compartments.
\item
H2: The rCBV is uniformly elevated around the outer rim of contrast enhancement, independent of our segmented NEH areas. If NEH regions represent biologically distinct, hypercellular tumor tissue, we would expect their rCBV values to be significantly higher than those of the peri-enhancing rims. Conversely, if elevated rCBV is simply a nonspecific artifact of the peri-enhancing region, no significant difference would be observed.
\item
H3: The preoperative NEH (pNEH) segment is expected to demonstrate spatial proximity to the site of recurrent contrast enhancement (ETRL), reflecting its biological relevance as a potential origin of tumor regrowth.
\eit

Each of these hypotheses is tested with one of the two datasets mentioned above.
\subsubsection{Testing H1 and H2 with the UPenn-GBM dataset}
First, we used the University of Pennsylvania Glioblastoma (UPenn-GBM) dataset, which provides a rich resource of imaging and clinical data not utilized during model training. This dataset includes $630$ patients with newly diagnosed glioblastoma, each with multi-parametric MRI scans (including T1, T1 with contrast (T1c), T2, FLAIR, diffusion tensor imaging (DTI), and dynamic susceptibility contrast (DSC) imaging) as well as comprehensive clinical information. Importantly, the availability of advanced DSC perfusion scans allows for the calculation of absolute pseudo-relative cerebral blood volume (ap-rCBV), a quantitative imaging biomarker that serves as a proxy for local cerebral blood volume.

It is well established that rCBV in brain tumors correlates with tumor cell density and is consequently elevated in non-enhancing hypercellular (NEH) regions compared to “pure” surrounding edema \cite{Sadeghi2008}, \cite{Hasanzadeh2023}. Based on this biological principle, we hypothesized that ap-rCBV could be used to validate the biological relevance of our segmented NEH regions.

To ensure robust and independent validation, we selected $390$ cases from the UPenn-GBM dataset, carefully excluding postoperative cases and any patients lacking complete perfusion or clinical data. Additionally, because $173$ cases in the UPenn-GBM dataset overlap with the BraTS 2018 dataset (which was used in model development), we excluded these overlapping cases to prevent any risk of data leakage and to guarantee that our validation cohort was entirely independent of the training data.

For each selected case, we calculated the NEH segment using the three network models described above. We then extracted the mean rCBV values from the predicted NEH, edema, and contrast-enhancing tumor regions by applying the calculated respective masks to the overall rCBV scans. These values were compared across the three regular compartments (necrosis, edema, enhancement) to test our hypothesis regarding the vascular and cellular characteristics of the NEH regions. This approach provides an objective, quantitative assessment of whether the model-defined NEH regions correspond to biologically meaningful, hypercellular tumor tissue as reflected by perfusion imaging.

To corroborate our findings and ensure that the observed increase of rCBV in NEH regions was not simply due to a general peri-enhancing artifact, we sought to test hypothesis H2. For this analysis, we used the segmentation mask of the contrast-enhancing tumor generated by the PAUNet model. It demonstrated a high Dice - S\o{}rensen score $(DSC = 0.8)$ for this compartment and showed strong agreement with other models in our study. We opted for this automated approach to maintain methodological consistency and efficiency, as repeating the process for all three models would be redundant given their high concordance for the enhancing tumor region.

Using the PAUNet-derived mask, we defined the outer rim of the enhancing tumor and then systematically inflated this rim outward by $2\ mm, 4\ mm$ and $6\ mm$ to create three concentric peritumoral regions. The selection of $2\ mm, 4\ mm$ and $6\ mm$ for the rim inflation distances was guided by the observed spatial characteristics of the NEH segments in our dataset. Specifically, when analyzing the segmented NEH regions across all cases, we found that their average thickness was approximately $4\ mm$. Since the NEH compartment typically forms a relatively thin layer surrounding the enhancing tumor, using rim extensions that span $2\ mm$ to $6\ mm$ allowed us to capture the spatial scale at which NEH regions most commonly occur. This approach ensures that our comparison between NEH and peri-enhancing rim regions is both anatomically relevant and tailored to the actual morphology of the NEH compartment observed in our study population. These inflated rims were then used to mask the co-registered rCBV maps, allowing us to calculate the mean rCBV values for each rim. To avoid confounding, we excluded necrotic areas, large vessels, and regions overlapping with the original enhancing tumor from the analysis. We then statistically compared the mean rCBV values of these peri-enhancing rims to those of the NEH segments.
It is important to note that elevated rCBV near the edges of enhancing tumors can result from several well-documented sources of artifact, including susceptibility effects, contrast agent leakage, and partial volume averaging at tissue boundaries \cite{Enzmann2006}, \cite{Dijken2019}, \cite{Patel2024}, \cite{Sugahara2001}, \cite{Martucci2023}. By explicitly comparing NEH and peri-enhancing rim regions, our approach aims to distinguish genuine biological differences from these potential imaging artifacts, thereby strengthening the validity of NEH as a distinct imaging biomarker in glioblastoma.

\subsubsection{Testing H3 using the RHUH glioblastoma dataset}
The next approach for clinical validation was to test hypothesis H3 by comparing preoperative NEH (pNEH) segments to the contrast-enhancing regions observed on MRI at first recurrence (ETRL). This approach is grounded in previous evidence showing a strong correlation between the proximity of preoperative hypercellular tumor regions (such as NEH) and the site of recurrence \cite{Kis2022}, \cite{Kim2021}. For this analysis, we utilized the Río Hortega University Hospital (RHUH) Glioblastoma dataset \cite{Kamnitsas2017}, which comprises longitudinal MRI scans-including pre-surgical, early post-surgical, and recurrence time points-for $40$ patients who all underwent either gross total or near-total tumor resection. This surgical homogeneity ensured that recurrence analysis was not confounded by substantial residual enhancing tumor.

Although the dataset includes automatic segmentations into three compartments (contrast enhancement, edema, necrosis) generated with Deep-Medic \cite{Avants2011}, we did not use these segmentations for our analysis, as they did not reliably distinguish the resection cavity from necrosis in recurrence scans. Instead, we selected $39$ patients (after excluding one with problems in the provided data sets) and performed our own segmentations:

\bit
\item
preoperative scans were segmented into four compartments (enhancing tumor (ET) compartment, nonenhancing hypercellular (NEH) compartment, edema (ED), necrosis (NC)) using PAUNET,
\item
recurrence scans were segmented into the same four compartments using UNETR++, which demonstrated the best performance for post-treatment imaging based on visual inspection.
\eit

It should be emphasized that our aim was not to establish a universally applicable methodology, but rather to provide proof-of-principle data supporting the clinical relevance of NEH segmentation in this specific context.

In detail, for spatial analysis, the preoperative non-enhancing segment (pNEH) was registered to the recurrence contrast-enhancing segment $(ETRL)$ using a combination of affine and non-rigid registration techniques with the ANTS program package \cite{Langhans2023}. Specifically, for the preoperative scans, masks included both the contrast-enhancing and necrotic regions, which together represented the typical target of surgical resection. In the recurrence scans, only the contrast-enhancing tumor was included in the mask. This approach ensured that the regions of interest were anatomically aligned and comparable across preoperative and recurrence imaging.

\subsubsection{Evaluation metrics}
To quantitatively assess the spatial relationship between the preoperative non-enhancing hypercellular (pNEH) segment and the contrast-enhancing region at recurrence ($ETRL$), we employed several spatial metrics and assessed spatial relationships between pNEH regions and recurrence sites ($ETRL$) by referencing established benchmarks for each metric.:
\bit
\item
{\em MeanEdgeDistance} measures the average distance from each voxel in the pNEH segment to the nearest edge of the $ETRL$ region, providing an overall sense of proximity between the two compartments. Values in the range of $1 \to 5\ mm$ are generally interpreted as indicating close anatomical proximity, with lower values reflecting a stronger spatial relationship between regions.
\item
{\em FractionNearEdge} calculates the proportion of pNEH voxels that are located within $5\ mm$ of the $ETRL$ boundary, highlighting the extent to which the pNEH segment is situated near the site of recurrence. Thresholds above $30\%$ are typically considered meaningful, suggesting that a substantial proportion of NEH voxels are located near the boundary of the recurrent tumor.
\item
{\em VolumeContainment} quantifies the absolute overlap volume between the pNEH and $ETRL$ regions. Values exceeding $10\%$ of the smaller region’s volume are often regarded as indicative of notable spatial overlap.
\item
{\em FractionInside} represents the fraction of the pNEH segment’s total volume that overlaps with the $ETRL$ region. Values above $20\%$ are interpreted as evidence of substantial containment of the NEH segment within the recurrence region.
\eit

These thresholds are informed by prior neuroimaging and tumor recurrence studies \cite{Liu2021}, \cite{Reinhardt1997}, \cite{Alexander2018}, \cite{Li2025}, and provide a framework for interpreting the observed spatial metrics in our analysis. Together, these metrics offer a comprehensive evaluation of both the proximity and spatial overlap between preoperative NEH regions and sites of tumor recurrence.

To assess whether the observed spatial metrics were significantly greater than expected by chance, a non-parametric permutation test was performed for each metric. This involved randomly flipping $10,000$ times the sign of deviations from a null baseline (e.g., zero for distances, one for ratios) and recalculating the mean value for each permutation. This generated a null distribution for each metric under the assumption of no systematic spatial relationship. The observed mean was then compared to this distribution to compute both one-sided and two-sided p-values.
\section{Results}

\begin{figure}[!htb]
 \begin{center}
 \includegraphics[width=0.7\textwidth]{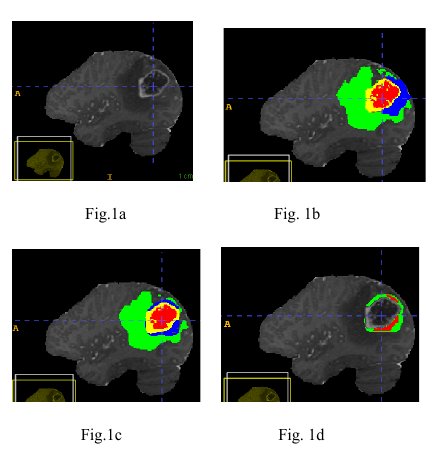}
\end{center}
 \caption{Subfigure 1a shows a plain T1 with contrast, 1b the PAUNET segmentation, 1c the UNETR++ segmentation, 1d UNETR-staple++ based segmentation. In the images with 4 segments ( 1b-c): green: edema, yellow: contrast enhancement, red: necrosis, blue: NEH. In 1d green denotes low probability, red denotes high probability}
 \label{fig:f1}
\end{figure}

Figure \ref{fig:f1} provides an example of segmentations achieved by the three nets. With these segmentations the further calculations and statistical analyses were performed.

\subsection{Quality assesment and feasibility of NEH segmentation}

Before presenting the clinical validation results, we first evaluated the segmentation performance of our models to establish the reliability of the predicted tumor compartments. We utilized the tumor segments as labeled in the BraTS 2021 competition. This dataset regularly contains four MRI modalities, namely T1c, T1, T2 and FLAIR. Accordingly Active Tumor (AT) comprises the enhancing tumor segment (ET = label 4), the TumorCore (TC) comprises the enhancing tumor (ET), the necrosis (NC) and nonenhancing tumor segments (NEH) (TC = labels 1,3,4) while the  Whole Tumor (WT) includes TumorCore plus edema (ED) (labels 1-4). For the calculations we combined the $309$ cases of the UPENN data set and $39$ preoperative cases from the RHUH data set into one data set resulting in $348$ cases in total. We compared the PAUNET segmentations to the ones of the UNETR++. Table \ref{tab:T1} shows the respective results for the S\o{}rensen-Dice coefficient, the Jacquard index, the Surface Dice and the Hausdorff 95 distance.

\begin{table}[!htb]
\caption{Comparison of PAUNET to UNETR++}
\label{tab:T1}
  \begin{center}
  \begin{tabular}{|l|l|l|l|}
   \hline
     Metric   & ET  & TC  & WT      \\
    \hline
    Dice & $0.83 \pm 0.15$ & $0.78 \pm 0.19$ & $0.90 \pm 0.07$ \\
    Hausdorff95 & $3.12 \pm 3.70$ & $12.43 \pm 17.82$ & $7.47 \pm 13.12$ \\
    Jacquard & $0.73 \pm 0.18$ & $0.68 \pm 0.22$ & $0.82 \pm 0.10$ \\
    Surface Dice & $0.85 \pm 0.15$ & $0.66 \pm 0.22$ & $0.76 \pm 0.12$ \\
    \hline
  \end{tabular}
  \end{center}
\end{table}

As can be seen, segmentation performance was highest for the whole tumor ($DSC(WT) = 0.90 \pm 0.07$), followed by the enhancing tumor ($DSC(ET) = 0.83 \pm 0.15$), and was lowest for the tumor core ($DSC(TC) = 0.78 \pm 0.19$). Boundary accuracy, as measured by the Hausdorff95 distance, was best for the enhancing tumor ($HD_{95}(ET) = 3.12 \pm 3.70\ mm$) but more variable for the tumor core ($HD_{95}(TC) = 12.43 \pm 17.82\ mm$) and whole tumor ($HD_{95}(WT) = 7.47 \pm 13.12\ mm$). Jacquard and Surface Dice metrics showed a similar pattern, with the highest spatial agreement for the whole tumor (WT) and the lowest for the tumor core (TC). These results are consistent with previous studies, reflecting the relative ease of segmenting the whole tumor and enhancing regions, and the greater challenge posed by the heterogeneous and ill-defined tumor core. Even recent models like DeepGlioSeg \cite{Bakas2017} report tumor core DSC scores of $DSC(TC) = 0.69 \to 0.75$ on BraTS datasets, highlighting persistent challenges even with advanced architectures.

Table \ref{tab:T2} present the metrics corresponding to the NEH tumor compartement separately (NC = label 3).

\begin{table}[!htb]
\caption{Average metrics for label 3 (NEH) achieved with PAUNet and UNETR++}
\label{tab:T2}
  \begin{center}
  \begin{tabular}{|l|l|}
   \hline
     Metric   & Label 3       \\
    \hline
    Dice & $0.29 \pm 0.19$ \\
    Hausdorff95 & $25.63 \pm 16.87$  \\
    Jacquard & $0.18 \pm 0.14$ \\
    Surface Dice & $0.40 \pm 0.20$ \\
    \hline
  \end{tabular}
  \end{center}
\end{table}

All metrics achieve low scores. These results are not unexpected, as NEH regions are known to be particularly difficult to segment. Our findings are consistent with literature reporting that non-enhancing or infiltrative tumor compartments are among the most challenging subregions for both automated algorithms and expert raters \cite{Vollmuth2025}.

Table \ref{tab:T3}, in addition, presents the results of the mean intensities for ED and NEH segments across the four MRI sequences. Here the differentiation between these two segments is crucial and difficult. Therefore it is important to show what the two nets achieve.

\begin{table}[!htb]
\caption{Mean intensities of areas masked as ED and NEH across the four MRI modalities achieved with PAUNet and UNETR++}
\label{tab:T3}
  \begin{center}
  \begin{tabular}{|l|l|l|l|}
  \hline
       &    & PAUNet & UNETR++ \\
   \hline
     Modality   & Label & mean intensity & mean intensity      \\
    \hline
    FLAIR & 2 & $374.62 \pm 101.68$  & $384.47 \pm 102.56$ \\
    FLAIR & 3 & $348.15 \pm 105.34$  & $362.55 \pm 104.19$ \\
    T1c   & 2 & $367.91 \pm 56.86$   & $362.57 \pm 55.75$ \\
    T1c   & 3 & $405.22 \pm 59.05$   & $390.11 \pm 57.45$ \\
    T1    & 2 & $340.63 \pm 55.62$   & $339.26 \pm 55.45$ \\
    T1    & 3 & $317.33 \pm 57.82$   & $317.47 \pm 56.36$ \\
    T2    & 2 & $570.71 \pm 267.36$  & $583.58 \pm 274.08$ \\
    T2    & 3 & $568.01 \pm 276.06$  & $583.40 \pm 288.11$ \\
    \hline
  \end{tabular}
  \end{center}
\end{table}

In order to test for significant differences we fitted a Linear Mixed-Effects Model. Table \ref{tab:T4} summarizes the results achieved with either PAUNet or UNETR++.

\begin{table}[!htb]
\caption{The statistics concerning the NEH compartment (label 3) obtained with PAUNet and UNETR++. The edema segment corresponds to label 2.}
\label{tab:T4}
  \begin{center}
  {\scriptsize
  \begin{tabular}{|l|l|l|l|l|l|l|}
  \hline
   \multicolumn{7}{|c|}{NEH tumor compartment - Label 3}  \\
   \hline
    Segmentation & Modality & Effect Estimate & t-value & F-value & Significance & Interpretation      \\
    \hline
    PAUNet  & FLAIR & $-26.55$ & $ -8.72$ & $ 76.03$ & $p \ll 0.001$ & $Int(L3) \ll Int(L2)$ by $~26.5$ units \\
    UNETR++ & FLAIR & $-21.74$ & $ -8.43$ & $ 71.08$ & $p \ll 0.001$ & $Int(L3) \ll Int(L2)$                  \\ \hline
    PAUNet  & T1c   & $+37.33$ & $+18.63$ & $346.95$ & $p \ll 0.001$ & $Int(L3) \gg Int(L2)$ by $~37.3$ units \\
    UNETR++ & T1c   & $+27.62$ & $+15.98$ & $255.30$ & $p \ll 0.001$ & $Int(L3) \gg Int(L2)$                  \\ \hline
    PAUNet  & T1    & $-23.28$ & $-14.72$ & $216.55$ & $p \ll 0.001$ & $Int(L3) \ll Int(L2)$ by $~23.3$ units \\
    UNETR++ & T1    & $-21.75$ & $-14.05$ & $197.45$ & $p \ll 0.001$ & $Int(L3) \ll Int(L2)$                  \\ \hline
    PAUNet  & T2    & $ -2.91$ & $ -0.62$ & $0.38$   & $p > 0.05$ & not significant \\
    UNETR++ & T2    & $ +0.16$ & $ 0.033$ & $0.0011$ &            & not significant \\
    \hline
  \end{tabular}
  }
  \end{center}
\end{table}

For both segmentation methods, the NEH tumor compartment (Label 3) exhibited lower intensity values than the edema compartment (Label 2) on FLAIR and T1-weighted images, while showing higher intensity on contrast-enhanced T1 (T1c). No significant difference was observed between the two labels on T2-weighted images. The differences in FLAIR, T1c, and T1 were substantial and reached high statistical significance with very small p-values. FLAIR results fit with typical radiological characteristics where edema is brighter than NEH, so seeing NEH intensity lower than edema intensity on FLAIR is plausible and expected. The T2 results concerning the NEH compartment are as different from those of the edema compartment as expected. The T2-weighted signal in NEH regions is generally hyperintense but often lower than that of pure edema due to higher cellularity. However, the distinction is not always pronounced, and NEH can sometimes be difficult to distinguish from edema on T2 alone, resulting in overlapping or variable signal intensities. The signal from the NEH compartment may sometimes appear higher in intensity than from the edema segment on T1c, not because of true enhancement, but due to higher cellularity or partial volume effects with adjacent enhancing tumor. In practice, the intensity difference between NEH and edema compartments on T1c cannot reliably be used for differentiation and can be neglected in our case \cite{Vollmuth2025}. From these data we can conclude that the proposed NEH segmentations correspond to the neuroradiological definitions \cite{Lasocki2017}.

\subsection{Clinical validation of segmentation methods}

\subsubsection{Comparison of rCBV values}
In a first step we compared the rCBV values of tumor compartments masked by the different models. We based this comparison on the UPENN dataset, which also contains rCBV data for each case. We masked the areas for edema (ED), contrast enhancement (ET), non-enhancing tumor (NEH) and necrosis (NC) utilizing the segmentation provided by the three models, PAUNet, UNETR++ and UNETR-staple++. We calculated the mean intensity of each mask and compared the resulting intensities by statistical tests. As the UNETR-staple++ yields two segments only, we calculated the masks of these high and low probability segments to obtain the mean magnitude and compare it to the other NEH masked rCBV data to check for plausibility.

\begin{figure}[!htb]
 \begin{center}
 \includegraphics[width=\textwidth]{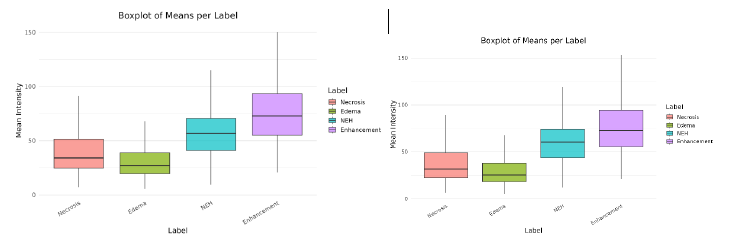}
\end{center}
 \caption{Mean CBV intensities by PAUNet and UNETR++}
 \label{fig:f2}
\end{figure}

Figure \ref{fig:f2} shows the masked rCBV levels for the four segmented regions obtained with the models PAUNet and UNETR++. The rCBV values are the highest for the enhancing tumor segment (ET), followed by the non-enhancing tumor segment (NEH) and the edema (ED). It is important to emphasize that the rCBV value for the edema compartment is lower than for the segmented NEH area. An ANOVA analysis yields a high significance $(p<0.0001)$ and the differences between the means are significant, too. This holds true for both U-Net models.

For the sake of completeness we included the necrosis segment (NC), though values are ambivalent depending on the intensity levels in it. But this is not relevant for our analysis.

Next, table \ref{tab:T7} shows the mean  rCBV levels of the NEH compartment achieved with PAUNet, UNETR++ and the UNETR-staple++ models. It shows that the UNETR-staple++ catches areas which are in correspondence with the two other network models. It shall be emphasized that the values with lower probability according to the UNETR-staple++ model are not significantly different from the high probability data.

\begin{table}[!htb]
\caption{Mean values of the rCBV intensity of the NEH compartment}
\label{tab:T7}
  \begin{center}
  \begin{tabular}{|l|l|}
   \hline
     Method   & NEH mean intensity      \\
    \hline
    PAUNet                & $57.39$ \\
    UNETR++               & $61.17$  \\
    UNETR-staple++(ll)    & $62.90$ \\
    UNETR-staple++(lh)    & $59.90$ \\
    UNETR-staple++(ll+lh) & $60.37$ \\
    \hline
  \end{tabular}
  \end{center}
\end{table}

All in all the rCBV data are consistent and correspond to reports from literature \cite{Sadeghi2008}, \cite{Hasanzadeh2023}. Furthermore the results strongly suggest that the NEH segments indeed have a biological meaning and do not simply depend on the mathematical procedure. Studies consistently show that rCBV is highest in the enhancing tumor compartment (ET) and decreases radially into peri-enhancing regions, reflecting a gradient from hypercellular tumor to edema/infiltrative margins \cite{Enzmann2006}, \cite{Dijken2019}.

\begin{figure}[!htb]
 \begin{center}
 \includegraphics[width=\textwidth]{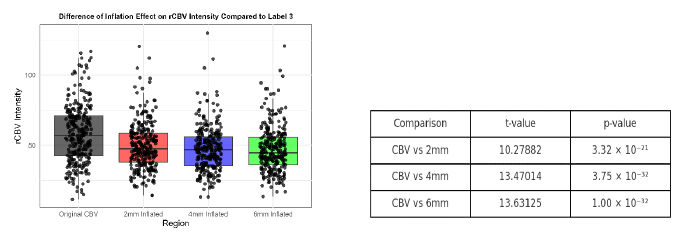}
\end{center}
 \caption{ Intensities of the different masked rCBV areas (left subimage). The segmented NEH segment is significantly higher than all the other inflated areas (right subimage)}
 \label{fig:f3}
\end{figure}

In order to make sure that we are not analyzing a well known artifact, we created masks of increasing thickness $(0 \to 2\ mm, 0 \to 4\ mm, 0 \to 6\ mm)$ around the outer rim of the enhancing tumor compartment (ET) and compared the corresponding means of the masked rCBV intensities. Figure \ref{fig:f3} illustrates the means and significances of the masked rCBV areas. One can clearly see that the segmented NEH tumor compartments exhibits the highest intensities. This difference is statistically significant $(p < 0.001)$ according to an ANOVA test. This result is quite important because it offers additional information. The interesting aspect to emphasize is that with increasing rim thickness the masks contain part or all of the NEH segment. As rim size increases $(0 \to 2\ mm \to 0 \to 6\ mm)$, the mean rCBV decreases because the rim incorporates lower-rCBV tissue (edema, normal parenchyma) that dilutes the NEH signal confirming that NEH represents a localized hypervascular compartment distinct from broader peri-enhancing effects. These additional results corroborate the validity of the separately segmented NEH, which would otherwise go unnoticed as merged in the peri-enhancing rings.

\subsubsection{Comparison of preoperative NEH to first recurrence}
It is well known that in more than $80 \%$ of the cases the recurrence of glioblastomas is located around the resection cavity of the primary operation. We wanted to check if we can find a spatial relation of the preoperative NEH segment (pNEH) to the enhancing tumor (ET) compartment at first recurrence. As explained above the Dice-S\o{}rensen coefficients and related metrics are not helpful in this context. We therefore chose local metrics for a spatial comparison. We considered $39$ cases of the RHUH data set for this analysis and  utilized the segmentation results of UNETR++ as basis. Label 4 of the preoperative dataset denotes the contrast enhancing tumor $(ET)_{re}$ compartment of first recurrence. For the evaluation, we computed the four spatial metrics as explained above.

\begin{figure}[!htb]
 \begin{center}
 \includegraphics[width=\textwidth]{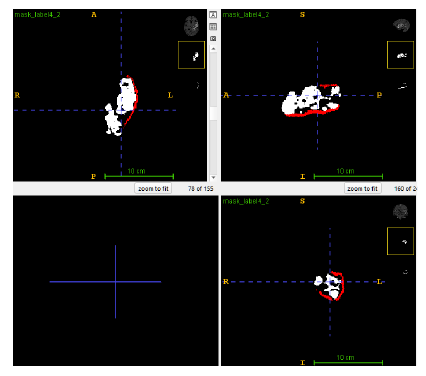}
\end{center}
 \caption{Preoperative NEH segment (red) warped on the recurrent ET compartment}
 \label{fig:f4}
\end{figure}

Figure \ref{fig:f4} exemplarily illustrates the matched segmentation for one case.

This figure just shows one possibility how the matching can occur. Part of the warped preoperative mask may also partially cover the remaining resection cavity. So variability is high and an additional statistical approach warranted.

\begin{table}[!htb]
\caption{Means and SD of the spatial metrics}
\label{tab:T8}
  \begin{center}
  \begin{tabular}{|l|l|}
   \hline
     Metric   & mean $\pm$ SD      \\
    \hline
    FractionInside        & $1.25 \pm 0.83$ \\
    FractionNearEdge      & $1.18 \pm 0.78$  \\
    MeanEdgeDistance      & $2.61 \pm 0.91$ \\
    VolumeContainment     & $3.63 \pm 1.31$ \\
    \hline
  \end{tabular}
  \end{center}
\end{table}

Table \ref{tab:T8} summarizes the means and standard deviations of the four spatial metrics used to evaluate the relationship between preoperative NEH regions and ET recurrence. The results indicate that, on average, NEH regions tend to be near and partially overlapping with ET recurrence areas, but not always in perfect alignment. More specifically,
\bit
\item
the {\em FractionInside} metric $(1.25 \pm 0.83)$ suggests that NEH regions slightly extend beyond the ET recurrence regions in many cases, though the high variability points to considerable inconsistency across patients. Similarly,
\item
the {\em FractionNearEdge} value $(1.18 \pm 0.78)$ shows that NEH regions are generally located close to the edge of ET recurrence zones, reinforcing the notion of spatial proximity, albeit not always showing precise overlap.
\item
The {\em MeanEdgeDistance} value $(2.61 \pm 0.91)$ further supports this observation, indicating a moderate average separation of approximately $2.6$ voxels between the edges of NEH and ET recurrence regions. Finally,
\item
the {\em VolumeContainment} metric $(3.63 \pm 1.31)$, reported on a log-transformed scale, reflects a broad spread in how much of the recurrence region is spatially contained within the NEH region, with some cases showing substantial overlap.
\eit

Taken together, these findings highlight a general spatial relationship between NEH and ET recurrence, while also emphasizing the significant inter-patient variability. This underlines the importance of individualized spatial analysis when interpreting NEH in the context of ET recurrence.

In order to go beyond a descriptive approach, we tested if the data represents results beyond pure chance. We consider this as relevant as the data base is not too big. We applied a non-parametric permutation test to each metric. The results are given in table \ref{tab:T9}.

\begin{table}[!htb]
\caption{Results of the permutation test for the spatial metrics.}
\label{tab:T9}
  \begin{center}
  \begin{tabular}{|l|l|l|l|l|}
   \hline
     Metric               & observed mean & null mean & one-sided p & two-sided p \\
    \hline
    FractionInside        & $1.251$       & $0.999$   &  0.0344     &  0.0719      \\
    FractionNearEdge      & $1.177$       & $0.998$   &  0.0820     &  0.1690      \\
    MeanEdgeDistance      & $2.613$       & $0.005$   & $<0.0001$     & $<0.0001$      \\
    VolumeContainment     & $3.627$       & $0.999$   & $<0.0001$     & $<0.0001$      \\
    \hline
  \end{tabular}
  \end{center}
\end{table}

The permutation tests revealed statistically significant deviations from chance for several spatial metrics:
\bit
\item
The {\em FractionInside} metric showed an observed mean of $1.251$ compared to a null mean of $0.999$, with a one-sided p-value of $0.034$, indicating that NEH regions tend to overlap with ET recurrence areas more than expected under random alignment.
\item
The {\em FractionNearEdge} value $(1.177 vs. 0.998)$ tended in the same direction but did not reach statistical significance $(p = 0.082)$ in a one-sided t-test.
\item
The {\em MeanEdgeDistance}, by contrast, showed a striking difference: the observed mean of $2.613$ was far above the null mean of $0.005$, with both one-sided and two-sided p-values $p < 0.0001$ indicating high significance.
\item
The {\em VolumeContainment}, similarly, showed a strong departure from chance given an observed mean: $3.627$ versus a null mean: $0.999$, also with high significance $p < 0.0001$ in both tests.
\eit

These results suggest a robust and non-random spatial relationship between preoperative NEH regions and areas of ET  recurrence. Permutation testing confirms that preoperative NEH regions exhibit statistically significant spatial association with ET recurrence, particularly in terms of edge proximity and volume overlap, indicating that these spatial patterns are unlikely to be due to chance. We emphasize that the approach applied in this study serves as proof of principle and is not intended as a predictive method for delineating future tumor recurrence. Rather, the goal is to demonstrate that preoperative NEH regions exhibit a statistically significant spatial relationship with ET
recurrence areas, thereby highlighting their potential biological and clinical relevance.
\section{Discussion}

Our results demonstrate that a mathematically deduced NEH segmentation, derived solely from routine clinical imaging, corresponds meaningfully with established biological and clinical markers. Specifically, we show that the NEH compartment — while non-enhancing on conventional imaging — exhibits rCBV characteristics consistent with hypercellular tumor tissue and demonstrates a statistically significant spatial relationship to future tumor recurrence. These findings provide compelling evidence that NEH segmentation identifies clinically relevant tumor components, despite the absence of a clear imaging gold standard.

\subsection{Biological interpretation}

The observed increase in rCBV values within the NEH regions supports the interpretation that these compartments represent biologically active, hypercellular tumor tissue. Previous work has established a link between rCBV and tumor cell density, microvascular proliferation, and tumor aggressiveness. Our results align with these findings and suggest that NEH regions may reflect the infiltrative edge of glioblastoma, which is often left unresected due to the absence of contrast enhancement. This is further supported by the spatial association between preoperative NEH segments and regions of contrast-enhancing tumor recurrence, suggesting that NEH may mark the zone of future tumor regrowth.

\subsection{Technical implications}

The use of a dual-network approach, combining convolution-based (PAUNet) and transformer-based (UNETR++) deep learning  models, contributes to the robustness and generalizability of NEH segmentation. PAUNet is optimized for fine-scale texture and edge detection, while UNETR++ incorporates global volumetric relationships via transformer encoders. Their complementary strengths were leveraged in a probabilistic consensus approach, further refined by STAPLE-like integration. This multi-model strategy is particularly well-suited to tasks like NEH segmentation, where boundaries are ill-defined and no single architectural paradigm suffices. The resulting probability maps are not only more stable but also biologically more plausible, as demonstrated by their correspondence with rCBV and ET recurrence patterns.

\subsection{Clinical and research applications}

While our method is not designed to predict tumor recurrence, it offers practical applications in both clinical and research contexts. NEH segmentation could support surgical planning by identifying non-enhancing tumor compartments that may benefit from supramarginal resection. In recent years, it has become increasingly evident that resection beyond the contrast-enhancing margin — commonly referred to as supramarginal resection — may improve outcomes by reducing residual infiltrative disease \cite{Acerbi2018}, \cite{Roh2023}, \cite{Petrecca2013}, \cite{Pessina2017}. However, one of the critical barriers to widespread adoption of this approach is the absence of a reliable, consensus-based method for defining the FLAIR-hyperintense margins typically targeted for resection \cite{Roh2023}.

Here, NEH segmentation may bridge this gap. By delineating biologically relevant, hypercellular, non-enhancing regions — regions shown in our study to correlate with both increased perfusion and spatial ET recurrence — our method offers a data-driven and reproducible target for guiding supramarginal resection beyond contrast enhancement. This could enhance surgical precision while reducing the subjectivity inherent in current FLAIR-based approaches.

Moreover, these segmentations could be integrated into clinical trial designs, allowing stratification based on infiltrative tumor burden or serving as imaging biomarkers to monitor treatment response in non-enhancing compartments. This is especially relevant given the growing interest in targeting the invasive margins of glioblastoma, where tumor  recurrence frequently originates.

\section{Limitations and Future Directions}

Several limitations must be acknowledged.
\bit
\item
First, our findings are based entirely on imaging data without direct histopathological correlation. Although rCBV and tumor recurrence provide strong biological validation, future studies, integrating biopsy or surgical specimens, would strengthen the clinical relevance of the NEH delineation.
\item
Second, our models were trained on datasets incorporating BraTS-derived segmentations, which may not fully represent the heterogeneity of glioblastoma across institutions.
\item
Third, while our spatial metrics and permutation testing provide rigorous support for a non-random relationship between NEH and ET recurrence, the sample size of the longitudinal dataset remains modest.
\eit

Future work should focus on prospective validation of NEH segmentation in surgical cohorts, integration with genomic and molecular markers, and longitudinal tracking of NEH evolution in response to therapy. Additionally, refinement of probabilistic thresholds and incorporation of uncertainty measures could improve clinical usability and interpretation.
\section{Summary}

In conclusion, our framework enables the automated, biologically grounded extraction of non-enhancing hypercellular tumor compartments using standard imaging inputs. While not intended to predict tumor recurrence, the statistically significant spatial and vascular associations observed here establish NEH as a reproducible and clinically relevant imaging feature. These results support the future use of NEH segmentation as a tool for individualized treatment planning, including decisions on supramarginal resection and personalized radiation targeting.

\section*{Acknowledgement}

Data were provided in part by the Human Connectome Project, WU-Minn Consortium (Principal Investigators: David Van Essen and Kamil Ugurbil; 1U54MH091657) funded by the 16 NIH Institutes and Centers that support the NIH Blueprint for Neuroscience Research; and by the McDonnell Center for Systems Neuroscience at Washington University.

\bibliographystyle{acm}
\bibliography{NEHSEG}
\end{document}